\documentclass[11pt]{amsart}
\pdfoutput=1
\usepackage{amsmath,amssymb,amsthm}
\usepackage{enumitem}
\usepackage{tikz}
\usepackage{tikz-cd}
\usetikzlibrary{decorations.pathmorphing}
\usepackage{bbm}
\usepackage{array}
\usepackage{hyperref}
\hypersetup{
    colorlinks=true,
    linkcolor=blue,
    filecolor=magenta,      
    urlcolor=blue}
\usepackage{xcolor}

\theoremstyle{plain}

\numberwithin{theorem}{section}

\theoremstyle{definition}

\newcommand{\el}{\ell}

\newtheorem*{rep@theorem}{\rep@title}
\newcommand{\newreptheorem}[2]{%
\newenvironment{rep#1}[1]{%
 \def\rep@title{#2 \ref{##1}}%
 \begin{rep@theorem}}%
 {\end{rep@theorem}}}

\newreptheorem{theorem}{Theorem}
\newreptheorem{lemma}{Lemma}
\newreptheorem{proposition}{Proposition}
\newreptheorem{definition}{Definition}
\newreptheorem{corollary}{Corollary}

\begin{document}



\centerline{\Large \sffamily Benchmarking PNW Model for MedMNIST to 100\% Accuracy}

\bigskip

\centerline{Bo Deng}

\centerline{\small Department of Mathematics}
\centerline{\small University of Nebraska-Lincoln}
\centerline{\small Lincoln, NE 68588}
\centerline{{\tt bdeng@math.unl.edu}}


\bigskip
\noindent \textbf{Abstract:} {In this paper, we introduce a new concept referred to as Artificial Special Intelligence by which Machine Learning models for classification problems can be trained error-free, thus acquiring the capability of not making repeating mistakes. The method is applied to 18 MedMNIST biomedical datasets. Except for three datasets, which suffer from the double-labeling problem, all are trained to perfection.}

\section{Introduction}
Artificial Special Intelligence (ASI) is defined as Artificial Neural Network (ANN) models in Machine Learning (ML) that do not repeat mistakes. Achieving ASI is training ML models error-free, which is both necessary and sufficient to avoid repeating errors. This requirement is 
self-evident for AI applications in healthcare, where devices known to repeat errors must not be accepted, 
for ethical principles, regulatory mandates, legal liability, and good care. It is also essential
for robots entering into domestic services that they must be capable of learning from mistakes under human supervision. In taxonomy, library catalog classification, or 
any data-based matching and prediction, where data accuracy and precision are supreme, ASI is a necessity. 

An ML classification problem is to construct a function or model that maps an input image to a label. For stand-alone artificial neural network (ANN) models, the Gradient Descent Tunneling (GDT) method can train them error-free (\cite{deng2023error-free}). However, when the training dataset becomes very large, say over fifty thousand, the GDT method works in theory, but becomes extremely slow and unreliable because of the large number of terms in the loss function for the gradient descent method.  

A model for data classification consists of two components: its architecture and labeling protocol. In this paper, we introduce a non-conventional model architecture and classification protocol that guarantees error-free training for training data of any finite size. Our model structure is motivated broadly by neuroscience in these aspects. First, contrary to the conventional deep neural network architecture, neural pathways from sensory modality to cognition are not deep. Second, the pathways are hierarchical, with the number of levels kept at a minimum. Third, the modal pathways are kept in parallel. Fourth, neurons or groups of neurons are specialized, firing action potentials only when excited by their modal stimuli. As for the classification protocol, we adopt a combination of majority voting and a winner-takes-all strategy, mimicking the neural states for cognition. 

As an illustration, the method is applied to tackle the image classification problem for 18 biomedical datasets from \cite{yang2023medmnist}. The benchmarking problem for these so-called MedMNIST datasets has been studied by researchers in multiple fronts, including a lightweight benchmarking in \cite{yang2023medmnist}, very deep neural network (DNN) models in  \cite{prvan2026lightweight}, foundational models in \cite{wu2025rethinking}, Convolutional Neural Networks (CNNs) and Vision Transformer (ViT) models in \cite{doerrich2025rethinking},  quantum computers approach with reduced resolution in \cite{singh2026benchmarking}, and some non-neural approach in \cite{karnes2026toward}. None of the methods reported in the literature achieved errorless training for any of the datasets. 
In contrast, our models achieve 100\% accuracy on 15 of the 18 datasets, as well as the three datasets when their double-labeled images are excluded.  

\section{Model Architecture}
\begin{figure}[t]
\centerline
{\scalebox{.5}{\includegraphics{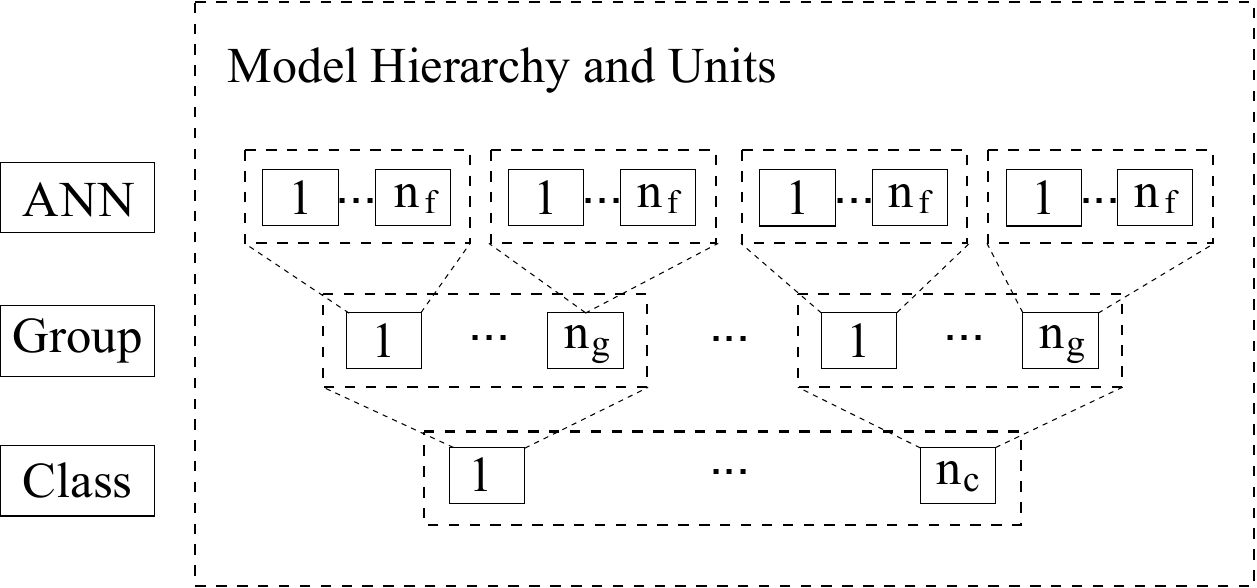}}}
\caption{PNW Architecture.}
 \label{fig_structure}
\end{figure}

As shown in Fig.\ref{fig_structure}, our model consists of three hierarchical processing units: ANN, Group, and Class, from low to high. At the base level, the model consists of a collection of ANNs, denoted by 
\[
A=\{A_{ijk}: 1\le i\le n_c,1\le j\le n_g,1\le k\le n_f\}
\]
where $n_c,n_g,n_f$ are natural numbers. Each ANN performs the task of classification, taking in an input datum and outputting a label from the label set
\[
L =\{1,2,\dots, n_\el, n_\el+1\}
\] 
for which $n_\el\ge 2$, and the $(n_\el+1)$th label is exclusively identified with an auxiliary label, referred to as the '\textit{expat}' label, which can also be read as 'not-from-my-class', or 'not-from-my-group'. 

The structure, referred to as a parallel neural web (PNW), is defined further as follows
\begin{enumerate}
    \item The first subscript of $A_{ijk}$ from the left is called a \textit{class}, and the model has $n_c$ many classes. If there is more than one class, then any two classes must share one and only one label, which is the auxiliary label expat. That is, the label set $L$ can be partitioned into $n_c$ subsets, $L=L_1\cup\cdots L_{n_c}$, so that images of class $C_i$ can be assigned labels only from $L_i$; and for two different classes, $C_i\ne C_t$, their label sets, $L_i$ and $L_t$, have only the expat label in common. 
    \item The second subscript from the left is called a \textit{group}, and the model has $n_g$ many groups for each class. All groups of a class can output the same labels of the class, including the expat label if $n_c\ge 2$. A group, $G_{ij}$, is identified by its class subscript $i$ and its group subscript $j$ for the $j$th group of class $C_i$.  
    \item The last subscript from the left is called a \textit{feature}, and the model has $n_f$ many features for each group. It can output the same labels of the group, including the expat label if $n_c\ge 2$.  
    \item A \textit{feature} of an input datum $x$ can be any transformation of $x$. We will use the word either to refer to the third subscript of an ANN, $A_{ijk}$, or to the transformation of $x$ that is used for the ANN, depending on the context, since little ambiguity can arise.
    \item The model architecture, therefore, consists of a collection of $N_T=n_c n_g n_f$ many ANNs, structured for three hierarchical levels from the features at the bottom to the classes at the top. 
    \item Every ANN is a processing unit. Every group of many ANNs is a processing unit. Every class of many groups is a processing unit. 
\end{enumerate}

 Examples of feature transformation, $x\to x_f$, include any binary black and white image of $x$, or any primary color channel of $x$, or any combination of the primary channels, or the edge image of $x$, or any transformation of $x$ from the field of Imaging Processing, such as Singular Value Decomposition transform, Fourier transform, Hough transformation of $x$, etc. The simplest feature for a gray-scale image is the original image, but in one vector form; for a color image, the three primary channels in one vector form; and for a 3D image, the 3D array in one vector form. 
 
In summary, ANNs are real structures, and each is trained on one feature of the input data. A feature is any transformation of the original input. In contrast, the group structure is somewhat virtual, by association mostly. That is, a group is a set of ANNs. If a group is a family, its ANNs are the members of the family. However, groups do differ when training is considered because different groups almost have no training data in common. The class structure is real because no two classes share ground-truth labels, except for the expat label. A class is a set of groups. All ANNs of a class share the same set of ground-truth labels, and two ANNs from two different classes have no common ground-truth labels. 

As a model, each group and each class, respectively, is a processing unit where the winning label of the model is determined. 

\section{Model Output Pathway}
\begin{figure}[t]
\vskip 1.0in
\centerline
{\scalebox{.5}{\includegraphics{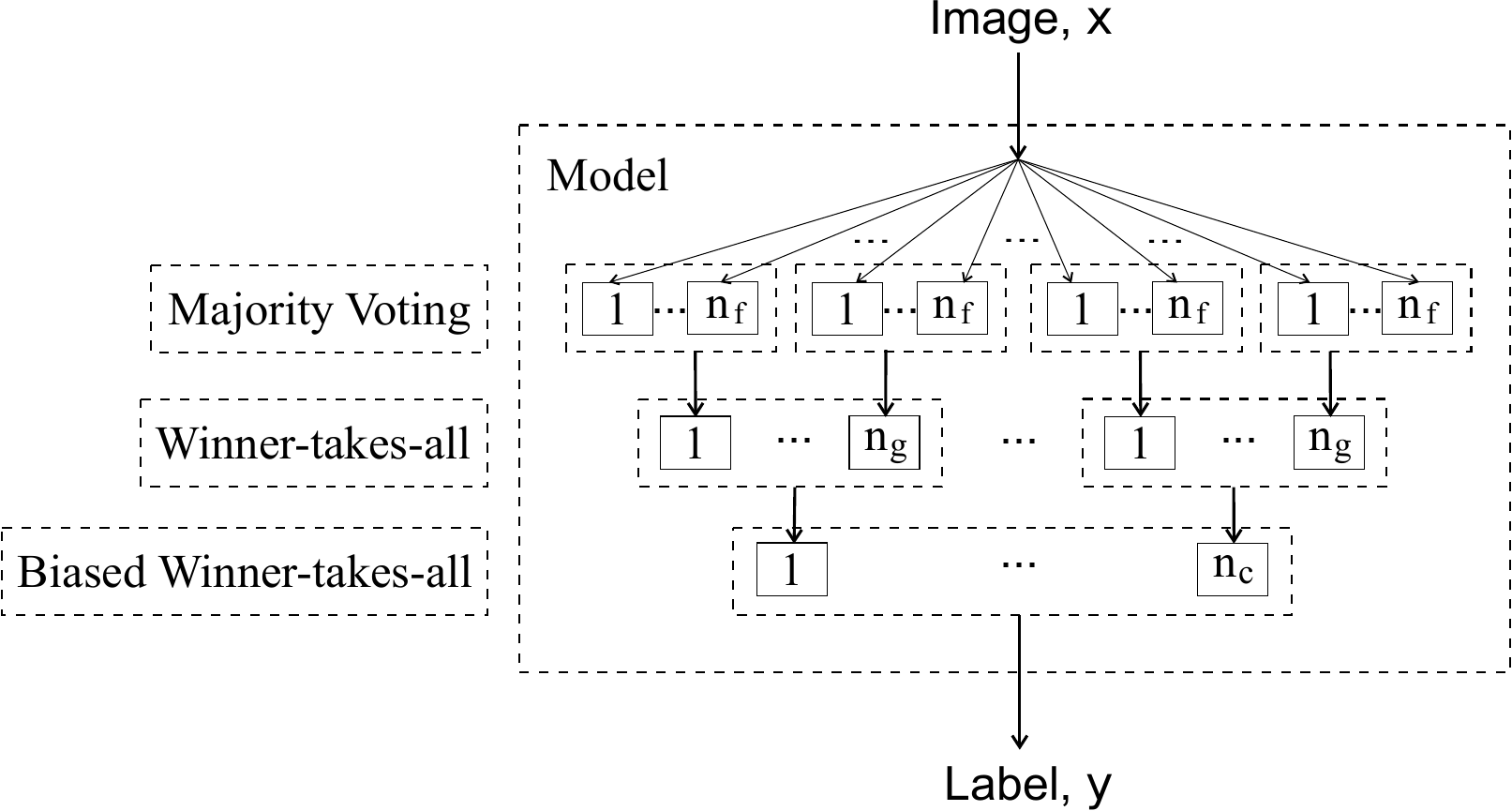}}}
\caption{Model Output Pathways.}
 \label{fig_model}
\end{figure}

As shown in Fig.\ref{fig_model}, for each datum input $x$, the output, $y=M(x)$, of the model is defined by an evaluation process through the hierarchy as follows. 
\begin{enumerate}
    \item  At the feature level, for every input, $x$, the ANN, $A_{ijk}$, outputs a label, $\el_{ijk}(x)$, from the label set $L$, and it also outputs the ANN's loss-function value of the label, $\varepsilon_{ijk}(x)$: 
    \[
        q_{ijk}=(\el_{ijk},\varepsilon_{ijk})=A_{ijk}(x),
    \]
    which is determined by the nearest neighbor protocol used by the  ANN for classification. 
    \item The number of output labels of every ANN is always greater than or equal to 2, and it always includes the auxiliary expat label if $n_c\ge 2$. Therefore, the number of classes is no greater than $n_\el$, the number of non-auxiliary labels, i.e., $1\le n_c\le n_\el$.
    \item For each group of a class, namely fixed $i$ and $j$ with $1\le i\le n_c$ and $1\le j\le n_g$, the evaluation process adopts a majority voting for the pool of labels, $\{\el_{ijk}: 1\le k\le n_f\}$, and outputs a label, the mode of the label (or the number of votes the label wins), and the winning label's loss-function value, denoted by 
    \[
    q_{ij}=(\el_{ij},v_{ij},\varepsilon_{ij}).
    \]
    Here $v_{ij}$ is the mode of $S=\{\el_{ijk}: 1\le k\le n_f\}$,  and $\el_{ij}$ is the label from the set $S$ with $v_{ij}$ many time(s). In the event of a tie, the smallest value of their loss-function values is used to break the tie. In the event that their loss-function values are identical, a uniform random choice from the tie labels is used to select the winning label, $\el_{ij}$, together with its loss-function value $\varepsilon_{ij}$. 
    \item For each class, i.e., a fixed $i$ with $1\le i\le n_c$, the evaluation process adopts a winner-takes-all protocol, outputting a label, its winning vote, and the winning label's loss-function value, denoted by 
    \[
    q_{i}=(\el_{i},v_{i},\varepsilon_{i}),
    \]
    where $v_{i}$ is the maximum of the vote subset $\{v_{ij}: 1\le j\le n_g\}$, and $\el_i$ is the corresponding label and $\varepsilon_{i}$ is the label's loss-function value. In the event of a tie, the minimum of their losses is used to break the tie. In the event that the losses are identical, a uniform random choice from the tie labels is used to break the tie. 
    \item Finally, for the model, the evaluation process adopts a biased winner-takes-all protocol, outputting a label, its winning vote, and the winning label's loss-function value, denoted by 
     \[
    q=(\el,v,\varepsilon),
    \]   
    where $v$ is the maximum of the vote subset $\{v_{i}: 1\le i\le n_c\}$, and $\varepsilon$ is the label's loss-function value. Unless all classes yield the expat label as their predictions, in which case, the outputting label is the expat label, and $v,\varepsilon$ can be assigned to the number of classes and 0, respectively, the winner-takes-all contest takes place among the non-expat labels only. The same tie-breaking protocol as above is used. And the output of the model, $y=M(x)$, is the winning label $y=\el$. 
\end{enumerate}

In summary, for any input datum, each ANN outputs a label, which is a vote for its group's output. A group's output is a label of the majority for the group. A class's output is the label having the largest group majority from its groups. Unless all classes output the expat label, the final output of the model is the label having the largest class majority from all the classes.  

\section{Training}
The purpose of introducing the class and group is to reduce the amount of training data each ANN uses to train itself. That amount for each ANN is referred to as the \textit{training load} of the ANN. The preliminary formation of the training data, $T_{ijk}$ for ANN $A_{ijk}$, is straightforward. The entire training dataset is first divided into classes of distinct label categories. For example, class 1 contains all animal images, and class 2 contains all plant images. Within each class, the load is further spread out among many groups. Roughly speaking, every training datum is assigned to one class, one group, and all featured ANNs of the group, referred to as its home group. All other groups are the datum's expat groups, and all other classes are its expat classes. A detailed description of the training subsets will appear elsewhere. 

Every ANN is then trained on its training data by the gradient descent method. First, by any stochastic gradient descent (SGD) method, e.g., \cite{shazeer2017outrageously}, followed by the gradient descent tunneling (GDT) method of \cite{deng2023error-free}. The GDT method can train each ANN error-free if the ANN's training data are free of the double-labeling problem, in which a datum appears at least two times but with different labels. Once all ANNs are trained to 100\% accuracy, the model's output protocol can follow. The model is considered fully trained if it correctly predicts every training datum. It can be proved that the PNW model can be fully trained by the method outlined above.     

\section{Result for MedMNIST}
As an example of application of our method, the MedMNIST datasets from \cite{yang2023medmnist} are considered. 
Table \ref{fig_model} summarizes the training result for all 18 datasets in the collection. In the Dataset column, replace * with MNIST to complete the dataset names. The first group, starting with `Breast*', is for 2D grayscale images. The second group is for 2D color images. The third group is for 3D grayscale images. The third column, `Label', is for the number of labels of a dataset. 

For the ChestMNIST, it has 14 label categories, each of which is a `Yes or No' questionnaire labeling. The training result is only for the first label category. The fourth, C-G-F, column is for the PNW model's architecture of a dataset, with C value for the number of classes, G for the number of groups, and F for the number of features. The fifth column is for individual ANN's architecture of the PNW models. The first number is for the image input size when vectorized. For example, all input sizes of 784 are from $28\times 28$ 2D grayscale images without transformation. All input sizes of 900 are for $64\times 64$ images, which are first down-sampled to $32\times 32$ and then have their borders trimmed off. The transformed images have a vectorized size of 900. 

For color images, each RGB channel results in a vector size of 900. Use the PathMNIST as an example, it has 18 featured ANN, each is trained on a cyclic R, G, B channel, for 6 cycles of the three channels. All input sizes of 2700 are for some featured ANNs for which the three transformed RGB channels are stacked into one vectorized input.

\begin{table}[t]
    \centering
    \begin{tabular}{l|rr|rr|rr}
Dataset & \multicolumn{2}{c|}{\text{Data Structure.}} & \multicolumn{2}{c|}{\text{Model PNW Arch.}} & \multicolumn{2}{c}{\text{ACC.}} \\ 
$\sim$MNIST & Set Size & Label & C-G-F & ANN & SGD & GDT \\ 
 \hline
\text{Breast*} &   546    &  2   &  1-1-3     &  900x125x2    &  100 & 100\\
\text{Chest*} &   78,468    &  \text{2(x14)}   &  \text{1-16-16}     &  \text{784x166x2}    &  97.012 & 99.995\\
&     \multicolumn{4}{r}{(without 10 double-labeled data)} & 97.016 & 100\\
\text{OCT*} &   97,477   &  4   &  2-10-20     &  784x116x3    &  96.937 & 97.019\\
&     \multicolumn{4}{r}{(without 2,906 double-labeled data)} & 99.725 & 100\\
\text{OrganA*} &  34,561    &  11   &  2-4-4     &  900x256x6/7   &  99.997 & 100\\
\text{OrganC*} &   12,975    &  11   &  1-1-2     &  900x125x11   &  100 & 100\\
\text{OrganS*} &   13,932    &  11   &  1-2-2     &  900x125x11   &  100 & 100\\
\text{Pneumonia*} &  4,708    &  2   &  1-1-3     &  900x125x2  & 100 & 100\\
\text{Tissue*} &  165,466    &  8   &  1-16-16  &  784x125x8    &  98.301 & 99.465\\
&     \multicolumn{4}{r}{(without 885 double-labeled data)}  & 98.826 & 100\\
\hline
\text{Blood*} &  11,959   &  8   &  2-3-6     &  2700x116x5    &  99.916 & 100\\
\text{Derma*} &  7,007  &  7   &  1-3-4     &  2700x256x7     &  99.986 & 100\\
\text{Path*} &  89,996   &  9   &  3-6-18     &  900x188x4    &  99.788 & 100\\
\text{Retina*} &  1,080    &  5   &  1-1-3     &  900x116x5    &  100 & 100\\
\hline
\text{Adrenal*3D} &  1,188    &  2   &  1-1-3     &  21952x125x2    &  100 & 100\\
\text{Fracture*3D} &  1,027    &  3   &  1-1-3     &  21952x125x3    &  100 & 100\\
\text{Nodule*3D} &  1,158    &  2   &  1-1-3     &  21952x116x2    &  100 & 100\\
\text{Organ*3D} &   971  &  11   &  1-1-3     &  21952x116x11    &  100 & 100\\
\text{Synapse*3D} &   1,230    &  2   &  1-1-3     &  21952x125x2    &  100 & 100\\
\text{Vessel*3D} &  1,335   &  2   &  1-1-3     &  21952x125x2    &  100 & 100\\

    \end{tabular}
    \vskip .2in
    \caption{Structure and accuracy for the MedMNIST PNW models.    }
    \label{tab:model_stat}
\end{table}

All input sizes of 21,952 are for the 3D images of size $28\times 28\times 28$ without transformation, resulting in the vectorized size of $28^3=21,952$. The second number of individual ANN's architecture is the number of nodes for their hidden layers. Various numbers are used. One can use a number anywhere in the range between 30 and 300 for any of the ANNs. The smaller the number, the longer it takes to train by the GDT method. The larger the number, the longer it takes to train because of the increased size. An optimal spot for the hidden-layer's nodes is around 100. All ANNs use one hidden layer. 

The last number of individual ANN's architecture is for the number of nodes of the classification layer. For PNW of a single class, the number is the same as the number of ground-truth labels. Otherwise, the number varies, which is the number of labels assigned to the class plus 1 for the expat label. Take the OrganAMNIST, for example. The first class of its PNW contains all images of the first 5 labels, for which their featured ANNs have $5+1=6$ labels for training. The second class contains all images of the last 6 labels, resulting in 7 classification output nodes for each ANN. 

The last two columns are for the accuracy of two types of trained PNW models. The first, SGD, is for the models trained only by our SGD method. For datasets of size over 10,000, error-free training is not achieved. When the SGD training is followed up by our GDT training, all datasets, free of the double-labeling problem, are trained to 100\% accuracy rate. For the three datasets suffering from the double-labeling problem, the double-labeled images are identified. For example, these paired image numbers,
\[
\{8936  ,     64078\},\ 
\{39439  ,     53059\},\ 
\{31396  ,     60808\},\ 
\{38396  ,     61196\},
\]
represent four images from the ChestMNIST dataset, each of which appears twice in the dataset, with different labels.      

\section{Discussion}
Errorless training for AI models represents a special kind of AI.
This capability is extremely important for AI applications in
healthcare. Using AI devices that have known operational errors
should be considered unethical. As a result, all biomedical image
classification models trained by conventional methods should not be
approved for clinical use. Some of the current regulatory standards, 
as outlined in \cite{USDA25}, are outdated because of the
technological advancements reported in this paper. In fact, as
part of patients' bill of rights in the future, medical AI assistants must be free of training errors and must be able to retrain to
eliminate future errors. Although all systems will make
mistakes, humans included, there is a fundamental difference
between the two types of errors; one type is of known-unknown, which may never happen, and the other type is of known-known, which has happened. Fully-trained AI assistants are of the first kind, and those with training errors are of the second kind. 

A preliminary version of the method was used for the dataset ImageNet-1k from \cite{ImageNet1kData}. Similar to ChestMNIST, OCTMNIST, and TissueMNIST, it is also plagued with the double-labeling problem, which prevents it from being trained error-free. We used a 20-2-17 PNW architecture for the dataset and achieved a model accuracy of 98.289\% (\cite{deng2025toward}
). For comparison, the MNIST dataset of handwritten digits (\cite{ MNISTData}) can be trained to zero error rate by the simple PNW model structure as shown in Tab.\ref{tab_model_stat} (\cite{deng2023error-free, Bench_MNIST}). All have set the benchmarks for other methods to meet. 

\begin{table}[h]
    \begin{tabular}{r|r|r|r}
\text{ANN Arch.} & \text{\# of Param.} & \text{Training Acc.} & Test Acc.\\
\hline
 784$\times 20\times$10    &  15,910 & 100\% & 94.26\% \\
 784$\times 40\times$10    &  31,810 & 100\% & 96.08\%  \\
 784$\times 60\times$10    &  47,710 & 100\% & 96.34\%  \\
 784$\times 80\times$10    &  63,610 & 100\% & 96.58\%  \\
 784$\times 100\times$10   &  79,510 & 100\% & 97.14\%  \\
    \end{tabular}
    \vskip .2in
    \caption{Benchmark for MNIST by PNW 1-1-1 Models}
    \label{tab_model_stat}
\end{table}

\begin{figure}[t]
\vskip 1.0in
\centerline
{\scalebox{.4}{\includegraphics{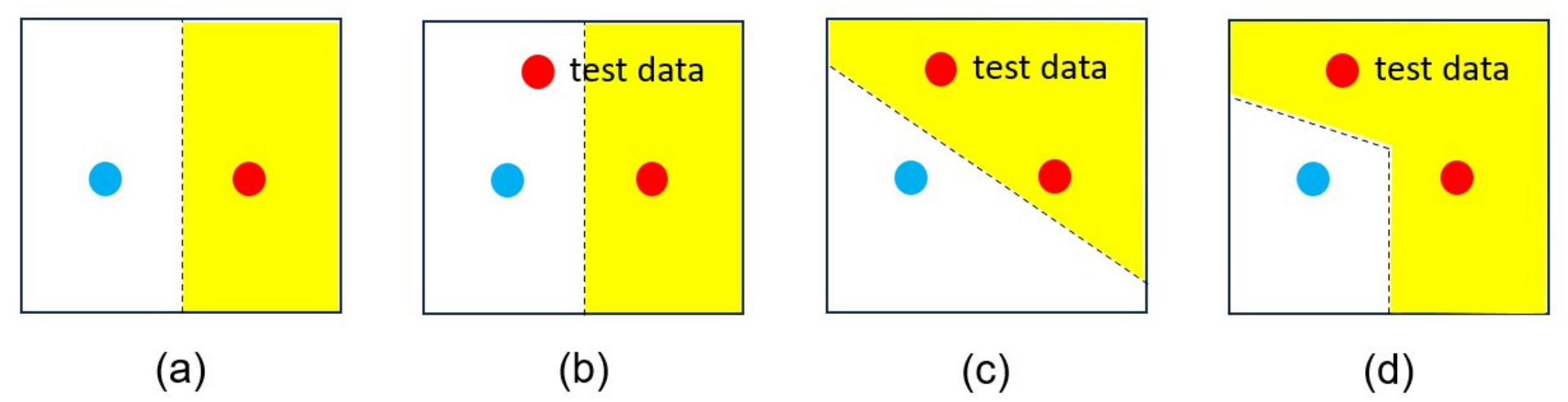}}}
\caption{No more test data: (a) Trained Voronoi cells without the test data. (b) The test data results in an error. Either (c) or (b) is a more accurate Voronoi cell formation than (a).}
\label{figVoronoi}
\end{figure}

Errorless training method brings about a paradigm shift for ML training. The first realization is that the data space is partitioned by error-free training algorithms into Voronoi cells, each containing some training data as seeds. The cell formation is the result of the nearest-neighbor protocol used for all classification problems. A prediction of a model is correct if the data lies in the cell to which it belongs. Otherwise, an error is made by the model. The higher accuracy that a training method achieves, the more accurate the trained Voronoi cells become. When the training method is capable of error-free, there is no reason to set aside any data for validation or test, because any data left behind for training will leave the Voronoi cell formation incomplete and incorrect with respect to the data on hand. This point can be simply demonstrated in Fig.\ref{figVoronoi}. As a result, we can conclude that all data must be included for training. In addition, we note that the concept of test data or validation data can not be properly defined. Our view is that what is not definable has limited use for theory and applications. 

As for the MedMNIST datasets, setting aside data for validation or testing does not serve any useful purpose because, regardless of anything we would do with them, for practical applications, all data must be trained before deployment. This was the reason that we did not consider the validation and test data from MedMNIST, and just used their training data for a demonstration of our method. 

\begin{figure}[t]
\vskip 1.0in
\centerline
{\scalebox{.5}{\includegraphics{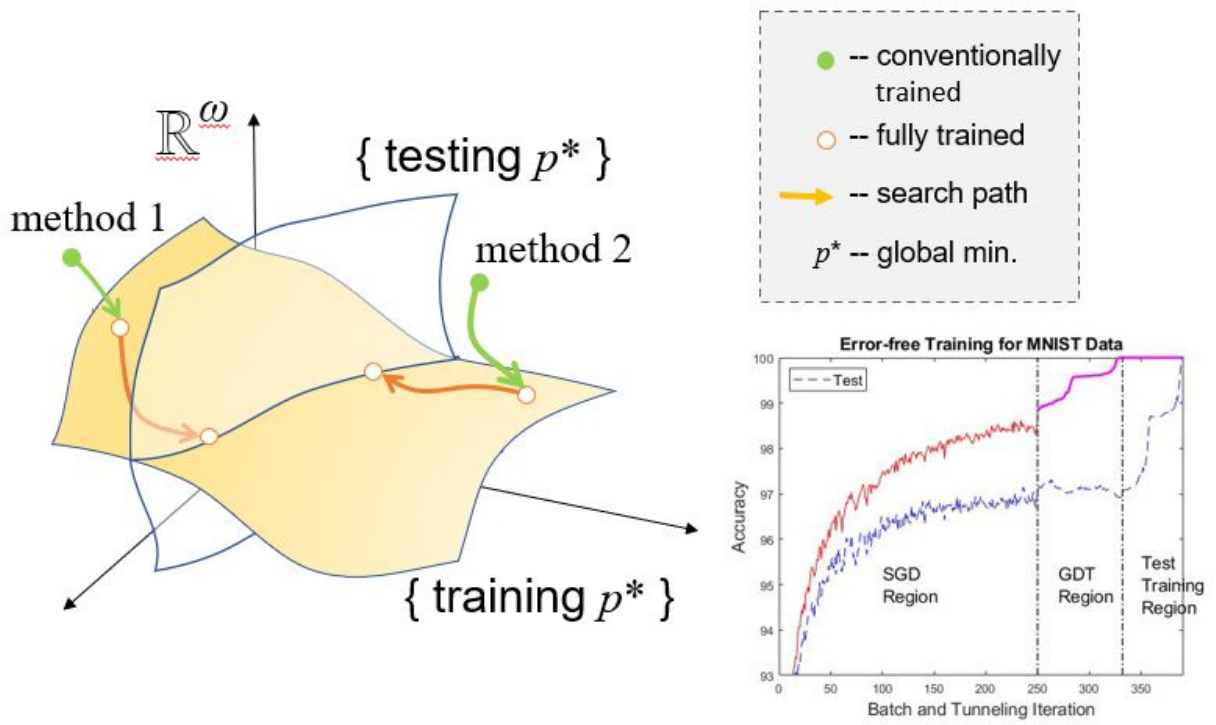}}}
\caption{Re-evaluation of Overfitting: The search path of 'method 2' on its way to the global minima would be misconstrued as an overfitting, while ignoring all other paths which accomplish the same training goal. The inserted plot is for a typical training path for the MNIST handwritten digits dataset from \cite{deng2023error-free}, which shows the training curves of the traditional SGD, the GDT for both the training data and then for the test data combined, whose training path ends at the joint global minimum sub-manifold for both the training and the test data.}
 \label{figGlobeMin}
\end{figure}

To train an ML model is to train it fully, i.e., to find the global minimum of the loss function. This is the only correct definition of training. When this is possible, there are only good fittings v.s.  bad fittings, by a training algorithm. As shown in Fig.\ref{figGlobeMin}, in the model parameter space, $\mathbb{R}^\omega$, the global minima consist of a sub-manifold, $\{{\rm training\ } p^\ast\}$, on which every point represents a fully trained model. If a set of data is set aside for test, as conventionally done, it too has a sub-manifold on which the test data are fully trained. Our errorless training algorithm guarantees the existence of both sub-manifolds, as well as their nonempty intersection, on which all data are fully trained. For parameter points off these manifolds, it corresponds to models achieving only sub-bar accuracies. Good fittings are the ones landing on the global minimum sub-manifold, $\{{\rm training\ } p^\ast\}$. Better fittings are the ones landing on the intersection, $\{{\rm training\ } p^\ast\}\cap \{{\rm testing\ } p^\ast\}$. Bad fittings are any points off these sub-manifolds. 

In the parameter space, a model's training is represented by a directional path, with the end point being the end of training. Overfitting is a type of search path that moves away first from the global minimum sub-manifold $\{{\rm testing\ } p^\ast\}$ before landing on the training global minima. It is just one possible type of infinitely many other path types along which the accuracy always increases. Overfitting is an illusion becoming an urban legend. It is often used to hide bad fittings.  

All AI training algorithms are based on the Universal Approximation Theorem (UAT) from \cite{Hornik1989,cybenko1989approximation}. The most important version of the UAT is the discrete version, which states that for any finite dataset, it can always be trained error-free by an ANN with only one hidden layer. The GDT method from \cite{deng2023error-free} is the first computation algorithm to realize the potential of the discrete UAT. The PNW architecture makes the GDT method universally applicable. That is, GDT-powered PNW can train any finite dataset error-free if the dataset is free of the double-labeling problem. We believe that our method is fundamental to ASI, and ASI is a fundamental integral part of the future of AI.  

\medskip\noindent
\textbf{Acknowledgment:} All large-scale computations for the paper were carried out in the Holland Computing Center of the University of Nebraska System. Thanks to Dr. Levi Heath for converting the MedMNIST datasets to Matlab's data file format. 

\smallskip\noindent
\textbf{Data Availability:} Due to the large dataset size, model parameters and evaluation codes are provided for only three PNW models in the figshare repository, \cite{figs_MedMNIST_25}. These are: OrganAMNIST for 2d grayscale images, PathMNIST for 2D color images, and NoduleMNIST3D for 3D images. For model parameters of the other MedMNISTs or questions, send emails to the author. 

\bibliographystyle{abbrv}
\bibliography{bibliography}

@misc{figs_MedMNIST_25,
author = {Deng, Bo},
license = {CC BY 4.0},
title = {{Error-free Training for MedMNIST Datasets}},
version = {1.0.0},
note = {\url{https://doi.org/10.6084/m9.figshare.32050326}}
}

@article{USDA25,
author = {U.S. Food and Drug Administration},
title = {{Artificial intelligence-enabled device software functions: Lifecycle management and marketing submission recommendations}},
journal={Tech. Rep.},
volume={FDA-2024-D-5255},
  number={},
  pages={},
year={2025},
organization={FDA},
note = {\url{https://www.fda.gov/regulatory-information/search-fda-guidance-documents/}}
}

@article{cybenko1989approximation,
  title={Approximation by superpositions of a sigmoidal function},
  author={Cybenko, George},
  journal={Mathematics of control, signals and systems},
  volume={2},
  number={4},
  pages={303--314},
  year={1989},
  publisher={Springer}
}

@inproceedings{ImageNet1kData,
  title={Imagenet: A large-scale hierarchical image database},
  author={Deng, Jia and Dong, Wei and Socher, Richard and Li, Li-Jia and Li, Kai and Fei-Fei, Li},
  booktitle={2009 IEEE conference on computer vision and pattern recognition},
  pages={248--255},
  year={2009},
  organization={Ieee}
}

@article{MNISTData,
  title={Gradient-based learning applied to document recognition},
  author={LeCun, Yann and Bottou, L{\'e}on and Bengio, Yoshua and Haffner, Patrick},
  journal={Proceedings of the IEEE},
  volume={86},
  number={11},
  pages={2278--2324},
  year={2002},
  publisher={Ieee}
}

@article{deng2023error-free,
  title={Error-free Training for Artificial Neural Network},
  author={Deng, Bo},
  journal={arXiv preprint arXiv:2312.16060},
  year={2023}
}

@article{shazeer2017outrageously,
  title={Outrageously large neural networks: The sparsely-gated mixture-of-experts layer},
  author={Shazeer, Noam and Mirhoseini, Azalia and Maziarz, Krzysztof and Davis, Andy and Le, Quoc and Hinton, Geoffrey and Dean, Jeff},
  journal={arXiv preprint arXiv:1701.06538},
  year={2017}
}

@article{Hornik1989,
  title={Multilayer feedforward networks are universal approximators},
  author={Hornik, K. and Stinchcombe, M. and White, H.},
  journal={Neural Networks},
  volume={2},
number={5},
  pages={359--366},
  year={1989}
}

@article{yang2023medmnist,
  title={Medmnist v2-a large-scale lightweight benchmark for 2d and 3d biomedical image classification},
  author={Yang, Jiancheng and Shi, Rui and Wei, Donglai and Liu, Zequan and Zhao, Lin and Ke, Bilian and Pfister, Hanspeter and Ni, Bingbing},
  journal={Scientific Data},
  volume={10},
  number={1},
  pages={41},
  year={2023},
  publisher={Nature Publishing Group UK London}
}

@article{deng2025toward,
  title={Toward Errorless Training ImageNet-1k},
  author={Deng, Bo and Heath, Levi},
  journal={arXiv preprint arXiv:2508.04941},
  year={2025}
}

@article{prvan2026lightweight,
  title={Lightweight Neural Network Ensemble Models for Medical Image Classification with MedMNIST Dataset},
  author={Prvan, Marina and Musi{\'c}, Josip and {\v{C}}oko, Duje and Kristi{\'c}, Ante},
  journal={Electronics},
  volume={15},
  number={7},
  pages={1470},
  year={2026},
  publisher={MDPI}
}

@article{wu2025rethinking,
  title={Rethinking foundation models for medical image classification through a benchmark study on medmnist},
  author={Wu, Fuping and Papiez, Bartlomiej W},
  journal={arXiv preprint arXiv:2501.14685},
  year={2025}
}

@article{singh2026benchmarking,
  title={Benchmarking MedMNIST dataset on real quantum hardware},
  author={Singh, Gurinder and Jin, Hongni and Merz Jr, Kenneth M},
  journal={Scientific Reports},
  year={2026},
  publisher={Nature Publishing Group UK London}
}

@article{karnes2026toward,
  title={Toward Aristotelian Medical Representations: Backpropagation-Free Layer-wise Analysis for Interpretable Generalized Metric Learning on MedMNIST},
  author={Karnes, Michael and Yilmaz, Alper},
  journal={arXiv preprint arXiv:2604.06017},
  year={2026}
}

@article{doerrich2025rethinking,
  title={Rethinking model prototyping through the MedMNIST+ dataset collection},
  author={Doerrich, Sebastian and Di Salvo, Francesco and Brockmann, Julius and Ledig, Christian},
  journal={Scientific reports},
  volume={15},
  number={1},
  pages={7669},
  year={2025},
  publisher={Nature Publishing Group UK London}
}

@misc{Bench_MNIST,
author = {Deng, Bo},
license = {CC-BY-NC-ND-4.0},
title = {{Validation for error-free ANN models on MNIST}},
version = {2.0.0},
year={2023},
note = {\url{https://doi.org/10.6084/m9.figshare.24328756}}
}

\end{document}